\documentclass{article}
\usepackage{spconf,amsmath,graphicx}
\usepackage{multirow, subfigure, setspace}

\title{Data Augmentation for End-to-end Code-switching Speech Recognition}
\name{Chenpeng Du\textsuperscript{1}, Hao Li\textsuperscript{1}, Yizhou Lu\textsuperscript{1}, Lan Wang\textsuperscript{2}, Yanmin Qian\textsuperscript{1}}
\address{
    \textsuperscript{1}
    MoE Key Lab of Artificial Intelligence\\
    SpeechLab, Department of Computer Science and Engineering\\
    AI Institute, Shanghai Jiao Tong University, Shanghai, China\\
    \textsuperscript{2}
    CAS Key Laboratory of Human-Machine Intelligence-Synergy Systems\\
    Shenzhen Institutes of Advanced Technology\\
     \small\texttt{\{duchenpeng,lh575526,luyizhou4,yanminqian\}@sjtu.edu.cn, lan.wang@siat.ac.cn}
}

\begin{document}

\maketitle
\begin{abstract}
  Training a code-switching end-to-end automatic speech recognition (ASR) model normally requires a large amount of data, while code-switching data is often limited.
  In this paper, three novel approaches are proposed for code-switching data augmentation. Specifically, they are audio splicing with the existing code-switching data, and TTS with new code-switching texts generated by word translation or word insertion.
  Our experiments on 200 hours Mandarin-English code-switching dataset show that all the three proposed approaches yield significant improvements on code-switching ASR individually. Moreover, all the proposed approaches can be combined with recent popular SpecAugment, and an addition gain can be obtained. WER is significantly reduced by relative 24.0\% compared to the system without any data augmentation, and still relative 13.0\% gain compared to the system with only SpecAugment.
\end{abstract}

\begin{figure*}[t]
  \centering
  \subfigure[Audio splicing]{
    \includegraphics[height=6.5cm]{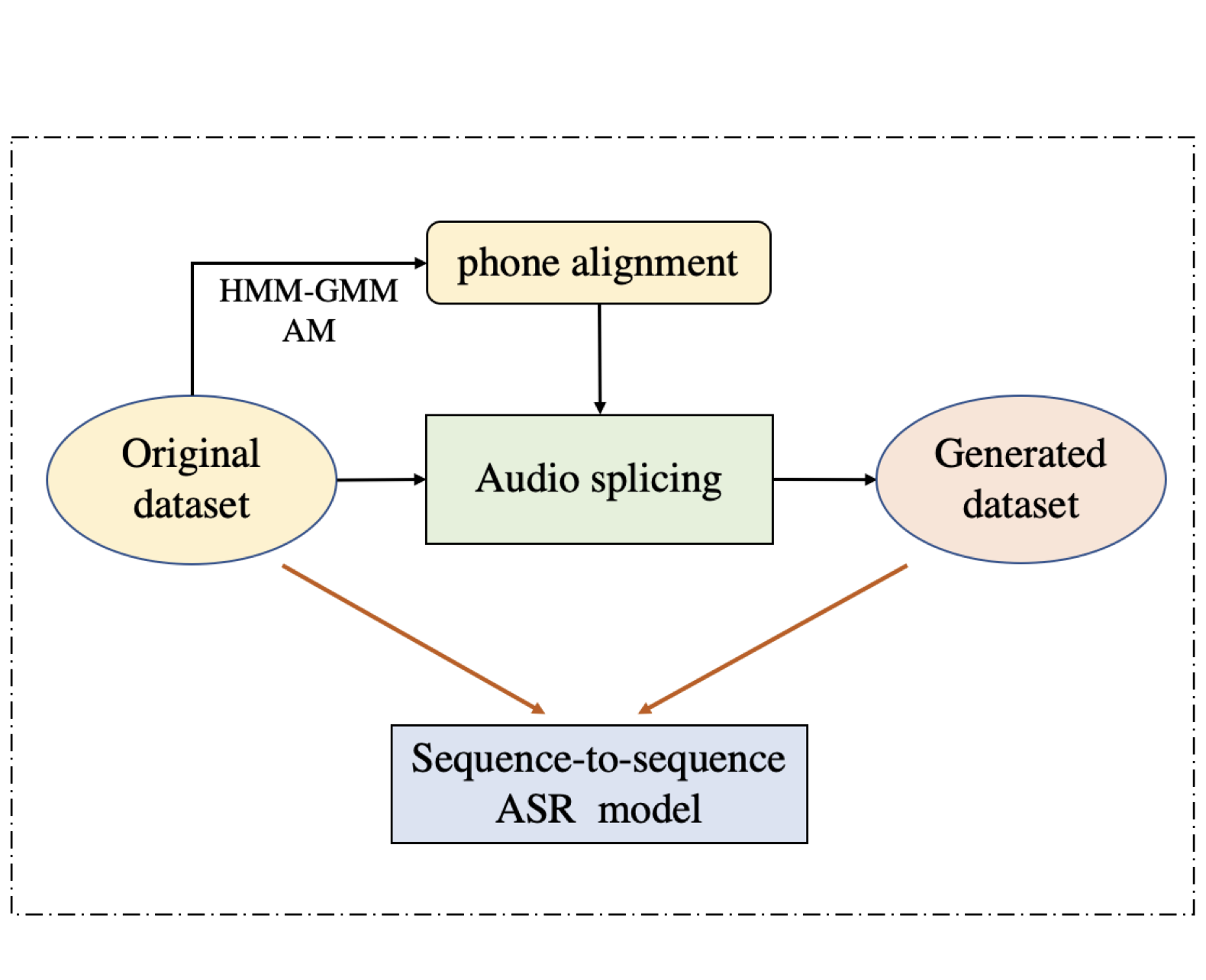}
    \label{process_as}
  }
  \hspace{3ex}
  \subfigure[TTS with word translation or insertion]{
    \includegraphics[height=6.5cm]{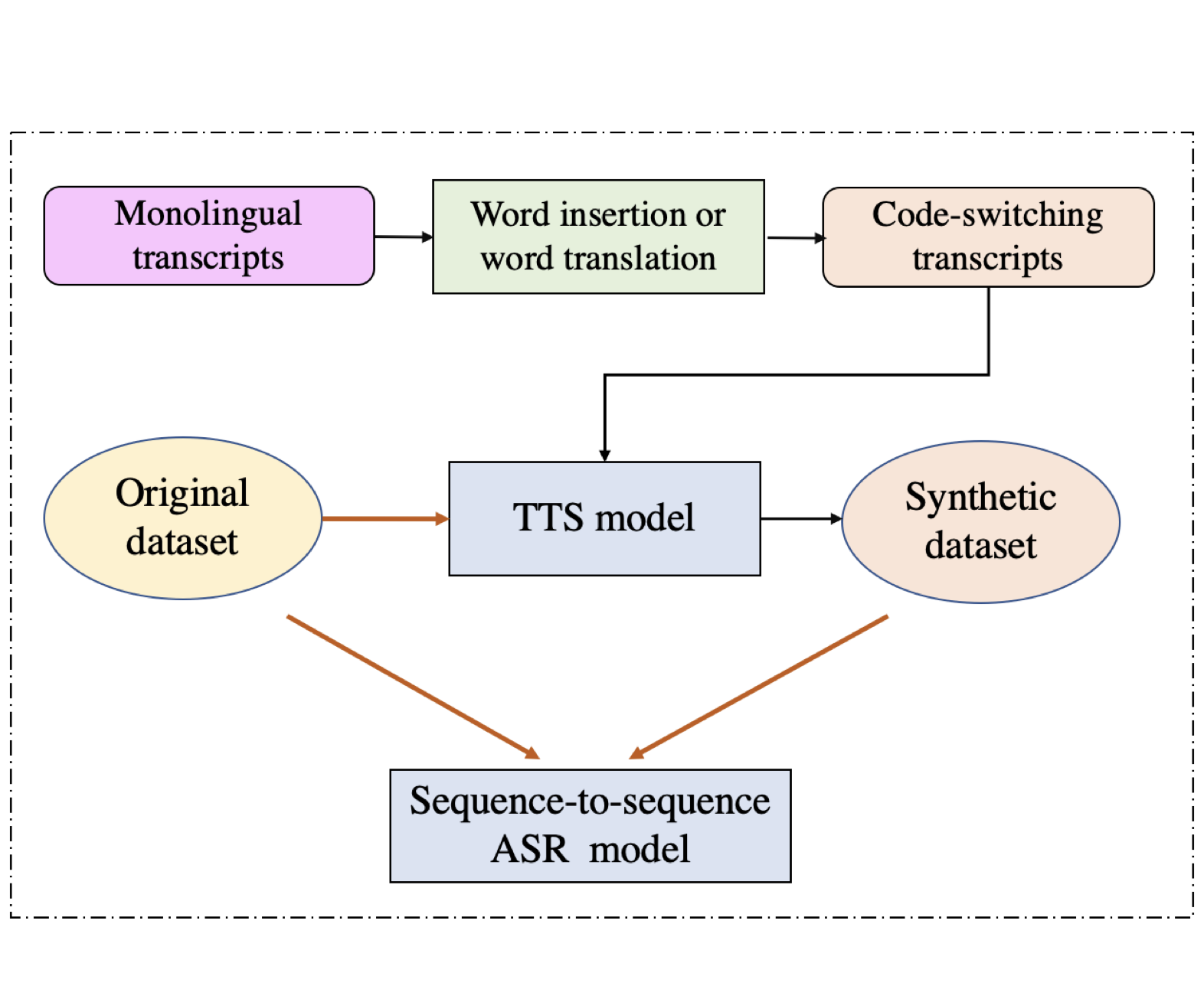}
    \label{process_tts}
  }
  \caption{Process of code-switching data augmentation methods proposed in this work}
  \label{fig:process_aug}
\end{figure*}

\section{Introduction}

Code-switching is a common language phenomenon that uses two or more languages within one utterance. For example, Chinese sometimes use English words in Mandarin conversations, which presents challenges to automatic speech recognition (ASR). The work in \cite{cs_boundary} gives an effective method to detect the language boundary in code-switching utterances and use monolingual ASR systems to recognize corresponding segments. \cite{bhuvanagiri2010approach, bhuvanagirir2012mixed} attempt to address the problem by constructing an appropriate pronunciation dictionary and modifying the language model. End-to-end ASR system for code-switching is also explored in recent studies \cite{e2e_cs_1, e2e_cs_2, e2e_cs_3, e2e_cs_4} that improves the performance of code-switching end-to-end ASR with language identity information. 

Typically, sufficient amount of training data is required for ASR training while code-switching data is difficult to be collected. Several approaches have been proposed to alleviate the problem. \cite{semi_1, semi_2} use semi-supervised training to leverage untranscribed data. \cite{khassanov2019constrained} encourages the distributions of output token embeddings of monolingual languages to be similar and trains code-switching ASR system with only monolingual data. \cite{cs_finetune} first trains the model using monolingual data and then finetunes on code-switching data.

Another important method to deal with this problem is data augmentation. Traditional data augmentation methods in ASR include noise augmentation \cite{addnoise}, speed perturbation \cite{speedperturb} and vocal tract length perturbation \cite{jaitly2013vocal}. Recently, SpecAugment \cite{park2019specaugment} is proposed as a powerful data augmentation approach using time warping and frequency masking. However, all these previous data augmentation approaches above only provide variations on acoustic features. 
\cite{concatutt} concatenates monolingual utterances from language-dependent corpora to generate code-switching data. However, this type of generated data only contains the code-switching that occurs between two utterances, while another type of code-switching that occurs within a single utterance is what we are interested in here.

In this paper, we propose three data augmentation methods for code-switching ASR training to improve both the acoustic and linguistic diversity of code-switching data. 
The first method is audio splicing. We splices language-dependent segments from various utterances together according to the alignments between transcripts and acoustic features, and thus generate new utterances for ASR training. 
Furthermore, although code-switching data is limited, monolingual texts is often easy to be collected, so we propose another two data augmentation methods to firstly generate new code-switching texts by word translation and word insertion on the original monolingual tests, and then produce new speech data using a TTS system. Experimental results show that all the three approaches yield improvements in code-switching ASR. Additionally, ASR still benefits from our approaches when SpecAugment is further combined.


In the rest of this paper, we introduce end-to-end ASR system in Section \ref{sec: asr} and the proposed code-switching data augmentation methods in Section \ref{sec: method}. Section \ref{sec: exp} gives experiments comparison and results analysis. Finally, we concludes the paper in Section \ref{sec: conclusion}.

\section{End-to-end Speech Recognition}
\label{sec: asr}

The end-to-end ASR system is briefly reviewed in this section. The neural network in E2E ASR models the conditional probability $p(\mathbf{y}|\mathbf{x})$, where $\mathbf{x}$ is a sequence of acoustic features and $\mathbf{y}$ is a sequence of labels. In this work, we use Transformer-based sequence-to-sequence (S2S) model \cite{transformer_asr_2, transformer_asr_4} jointly trained with CTC \cite{ctc, CTC_e2e}. Specifically, the encoder contains a front-end for subsampling and layers of Transformer blocks \cite{transformer}. The input $\mathbf{x}$ is passed through the encoder and mapped into a high level representation $\mathbf{h}$.

\begin{equation}
\begin{aligned}
\label{eq:encoder}
\mathbf{h} = Encoder(\mathbf{x})
\end{aligned}
\end{equation}

The decoder contains layers of Transformer blocks, a linear projection layer and a Softmax layer. Here, the Transformer blocks apply two attention mechanisms, i.e. self attention and the attention with encoder output $\mathbf{h}$. The posterior distribution of the next token is predicted with auto-regressive decoding.

\begin{equation}
\begin{aligned}
\label{eq:decoder}
p_{s2s}(\mathbf{y}_t;\mathbf{y}_1...\mathbf{y}_{t-1}) = Decoder(\mathbf{h},\mathbf{y}_1...\mathbf{y}_{t-1})
\end{aligned}
\end{equation}

Encoder output $\mathbf{h}$ is also passed through another linear projection layer and Softmax layer for CTC. We jointly train the S2S ASR system with CTC, and the loss function is formulated as:
\begin{equation}
\begin{aligned}
\label{eq:asr_loss}
L_{ASR} = -\alpha \cdot \log p_{s2s}(\mathbf{y};\mathbf{x}) - (1-\alpha) \cdot \log p_{ctc}(\mathbf{y};\mathbf{x})
\end{aligned}
\end{equation}
where $\alpha$ is a hyper-parameter to tune the relative weights between S2S and CTC losses.
In the decoding stage, we predict the output tokens with beam search, which combines the scores of S2S and CTC \cite{watanabe2017hybrid}.

\begin{figure*}[t]
  \centering

  \subfigure[Audio splicing]{
    \includegraphics[height=3.5cm]{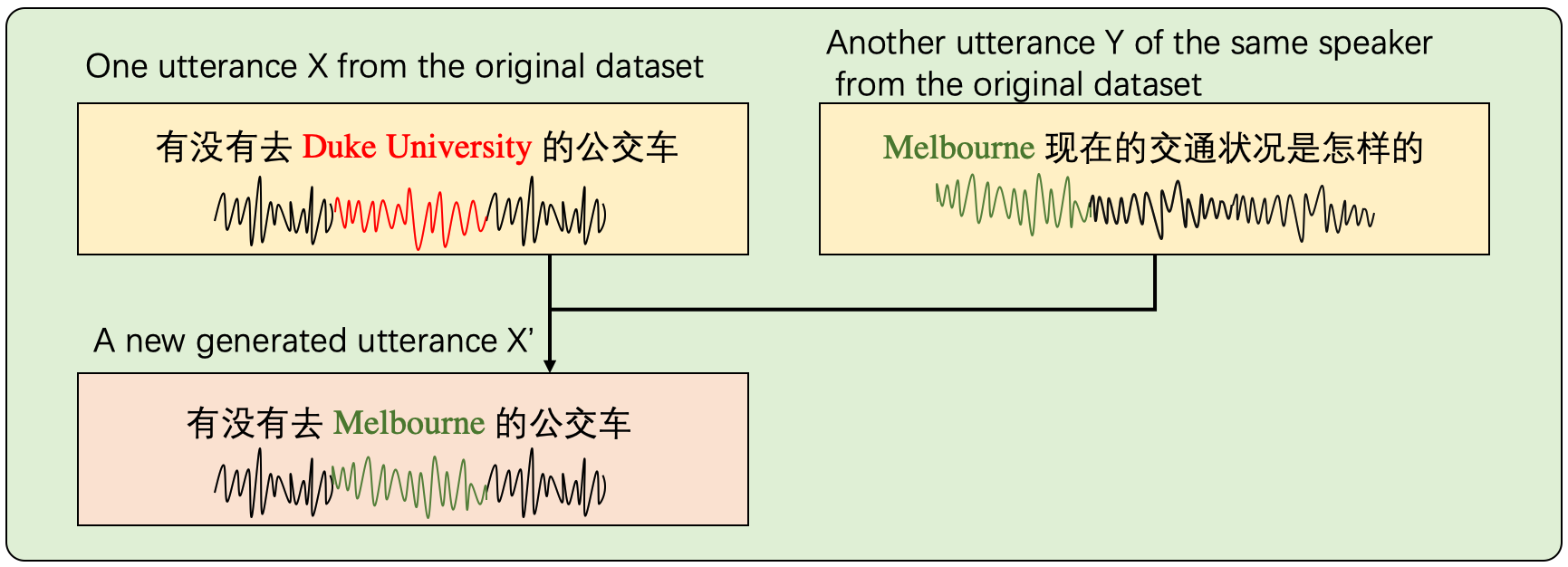}
    \label{audio_splicing_egs}
  }
  \subfigure[Word translation]{
    \includegraphics[height=3.5cm]{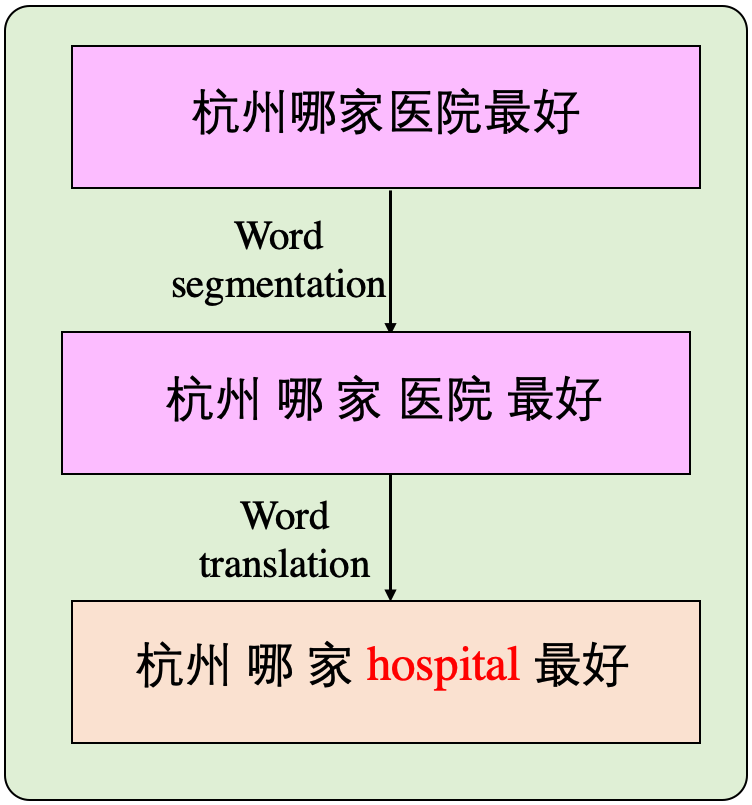}
    \label{word_trans_egs}
  }
  \subfigure[Word insertion]{
    \includegraphics[height=3.5cm]{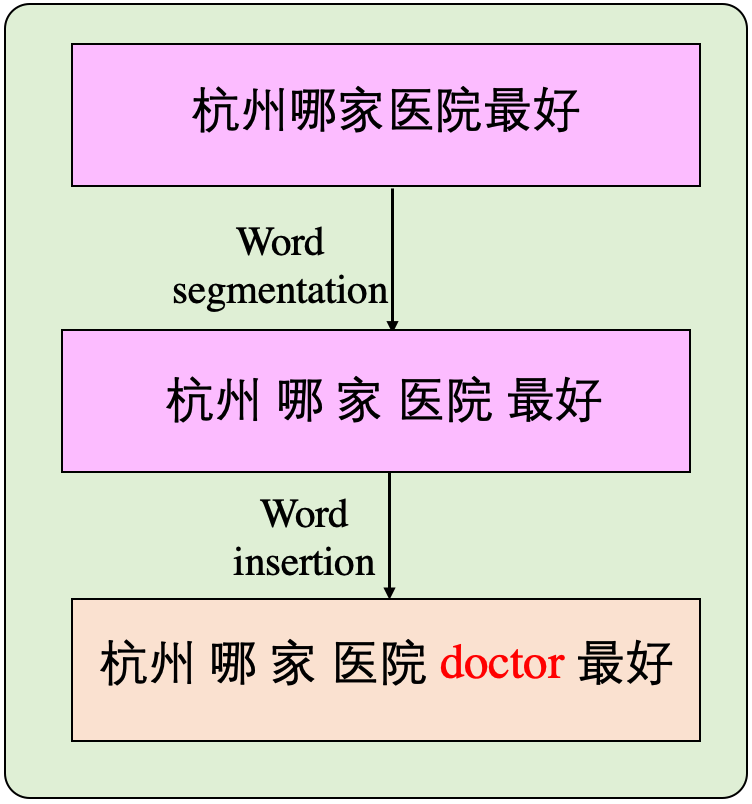}
    \label{word_ins_egs}
  }
  \caption{Augmented examples of the proposed code-switching data augmentation approaches}
  \label{fig:audio_splicing_method}
\end{figure*}

\section{Data augmentation for code-switching ASR}
\label{sec: method}

In this section, three novel data augmentation methods are developed for code-switching ASR, which can increase the data diversity on both acoustic and linguistic aspects.
\subsection{Audio splicing}
\label{sec: audio_splicing_method}

The first method is named audio splicing, and the process of data augmentation with audio splicing is illustrated in Figure \ref{process_as}. The original dataset contains code-switching utterances. We first train an HMM-GMM ASR system on the original dataset and obtain alignments on these code-switching data. According to the alignments, we can cut the individual language-dependent segments from the code-switching utterances. Then, we splice the segments randomly from different original utterances with the same speaker, and generate a new code-switching dataset. Both the original and the newly generated code-switching dataset are then pooled together to train the end-to-end ASR. 

 Figure \ref{audio_splicing_egs} illustrates an example of audio splicing in Mandarin-English. Specifically, for each utterance $\mathbf{X}$ in the original dataset, we randomly select another utterance $\mathbf{Y}$ of the same speaker. Then we replace the English segment of $\mathbf{X}$ with that of $\mathbf{Y}$, and thus generate a new code-switching utterance $\mathbf{X'}$, which will be used in the later ASR model training. 
 
\subsection{Code-switching text with word translation}
\label{sec: word_trans_method}

Code-switching data is limited on both speech and language aspects. Compared to code-switching text, monolingual text is much easier to be collected. To better utilize the monolingual data for code-switching task, we proposed new methods to change the monolingual text to new code-switching text. The first one is named word translation, and Figure \ref{word_trans_egs} illustrates an example of generating new code-switching text on Mandarin-English task. Specifically, we first conduct word segmentation and POS tagging on additional Mandarin transcripts. For each transcript, we then randomly select a noun or verb and conduct word-to-word translation with a Mandarin-English dictionary. Consequently, we obtain a new text corpus with code-switching transcripts. These new texts will be then passed to TTS system, which will be described in Section \ref{sec: tts_method}.

\subsection{Code-switching text with word insertion}
\label{sec: word_ins_method}
The second method is named word insertion. We could alternatively use word insertion to transform monolingual texts into code-switching ones. An example of word insertion is illustrated in Figure \ref{word_ins_egs}. Specifically, we conduct word segmentation on additional Mandarin transcripts. For each utterance, we then randomly select an English word from an English lexicon, and insert the word into the utterance at a random position. Similarly, we obtain a new text corpus with code-switching transcripts, which will be then passed to TTS system.

\subsection{Text-to-speech with new code-switching texts}
\label{sec: tts_method}

With these new generated code-switching texts, a TTS system is performed to generate the speech, and the complete process of TTS with word translation or insertion is shown in Figure \ref{process_tts}. Provided with the code-switching transcripts from word translation or word insertion, we synthesize the corresponding speech with a TTS model. Then the synthetic dataset and the original dataset are pooled together for end-to-end ASR training. 

Here, the TTS model is based on Fastspeech \cite{ren2019fastspeech} architecture and is trained with the original code-switching dataset. 
Duration targets $\mathbf{d}^{t}$ of standard Fastspeech are extracted from the attentions of an encoder-attention-decoder TTS model. However, we find that the attention alignments are sometimes poor. Therefore, in order to improve the quality and stability of synthetic speech, we train our Fastspeech-based TTS system with duration targets $\mathbf{d}^{t}$ from the alignments of an HMM-GMM ASR system.

Instead of using vectors from a speaker embedding table to indicate speakers, we sample speaker embeddings from a latent variable $\mathbf{z}$ of a variational auto-encoder (VAE) \cite{vae}. This is designed to improve the speaker diversity in synthetic speech \cite{speaker_aug}.
Prior distribution of $\mathbf{z}$ is set to an isotropic standard Gaussian $N(\mathbf{0},\mathbf{I})$. Our audio encoder architecture is similar to \cite{gst}, which takes mel spectrogram $\mathbf{y}$ as input and outputs the mean and log variance of the posterior probability distribution of latent variables. $\mathbf{z}$ is sampled from the posterior distribution $q(\mathbf{z}|\mathbf{y})$ in training and from the prior distribution $p(\mathbf{z})$ in generation. 

The loss function for TTS training can be formulated as
\begin{equation}
\begin{aligned}
\label{eq:loss}
L_{TTS} = &L_{MSE}(\mathbf{d}, \mathbf{d}^{t}) \\
&+ \mathbf{E}_{q(\mathbf{z}|\mathbf{y})}[\log p(\mathbf{y};\mathbf{x}, \mathbf{z})] \\
&+ \lambda\cdot D_{KL}[q(\mathbf{z}|\mathbf{y})||N(\mathbf{0},\mathbf{I})] 
\end{aligned}
\end{equation}
where $\lambda$ is the hyper-parameter to tune the relative weights. The first term denotes the MSE loss between predicted and extracted durations. The second term denotes the reconstruction loss between generated mel spectrograms and corresponding targets. The third term is KL-divergence between $q(\mathbf{z}|\mathbf{y})$ and $N(\mathbf{0},\mathbf{I})$. We use Griffin-Lim algorithm \cite{gl} to reconstruct waveforms from predicted Mel spectrograms.

\section{Experiment and Result}
\label{sec: exp}

\subsection{Dataset}
The dataset in ASRU 2019 Code-Switching ASR Challenge is used in our experiments, and it contains 200 hours Mandarin-English code-switching data and 500 hours Mandarin data, denoted as ASRU-CS and ASRU-MANDARIN here. Our experiments are mainly based on ASRU-CS, in which Mandarin is the host language and English is the guest language. The audio of ASRU-CS is recorded in a relatively quiet indoor environment with 16kHz sampling rate. Additional 20 hours Mandarin-English data is used for the evaluation, and we divide it into Mandarin part CER, English part WER and the total mix error rate (MER) as those in ASRU2019 Challenge.

\subsection{Baseline systems}

We use ESPnet \cite{watanabe2018espnet} toolkit to train our end-to-end ASR systems. 3003 Mandarin characters and 1000 English byte pair encodings (BPE) \cite{bpe} are used as modeling units. We set $\alpha$ in equation \ref{eq:asr_loss} to $0.2$ and use the same learning rate scheduling as in \cite{asr_warmup} for all experiments with initial learning rate of 5.0. We use Adam optimizer with $\beta_1=0.9$, $\beta_2=0.98$. 

\begin{table}[htbp]
\caption{WER (\%) on Mandarin-English test dataset of baseline systems}
\label{tab:baseline}
\centering
\begin{tabular}{l|ccc}
\hline
 \multirow{2}{*}{\textbf{Data Augmentation}}    & \multicolumn{3}{c}{\textbf{WER}}  \\
\cline{2-4} 
      & \textbf{CN}            & \textbf{EN}             & \textbf{TOTAL}                                 \\ 
\hline
 None    & 11.15         & 33.31          & 13.56         \\     
\hline
 Speed Perturb            & 10.86         & 32.77          & 13.23          \\   
  Monolingual TTS         & 11.13          & 31.61           & 13.35  \\  
 SpecAug               & 9.60          & 30.18          & 11.84          \\
\hline
\end{tabular}
\end{table}

The baseline system without data augmentation is trained on ASRU-CS. Moreover, we also implement some usual data augmentations to improve the CS ASR, and three basic methods are compared, including speech perturb, momolingual TTS and SpecAug. The speed perturbation uses speed ratios 0.9, 1.0 and 1.1. In the monolingual TTS method, we synthesize 200k monolingual utterances with 100k Mandarin transcripts from ASRU-MANDARIN and 100k English transcripts from Librispeech. Finally, the recent popular SpecAugment \cite{park2019specaugment} is also performed. The frequency mask parameter $F$ and the time mask parameter $T$ of SpecAugment is set to 30 and 40 respectively, the number of frequency masks $m_F$ and time masks $m_T$ are both set to 2, and the time warp parameter $W$ is set to 5.

As shown in Table \ref{tab:baseline}, only very limited gains are obtained by speed perturbation or monolingual TTS for this code-switching ASR task, and in contrast SpecAugment can yield significant improvement compared with the baseline.

\subsection{Evaluations on the proposed data-aug methods}

\begin{figure}[t]
\centering
\subfigure[Original data]{
\includegraphics[width=\linewidth,height=1.5cm]{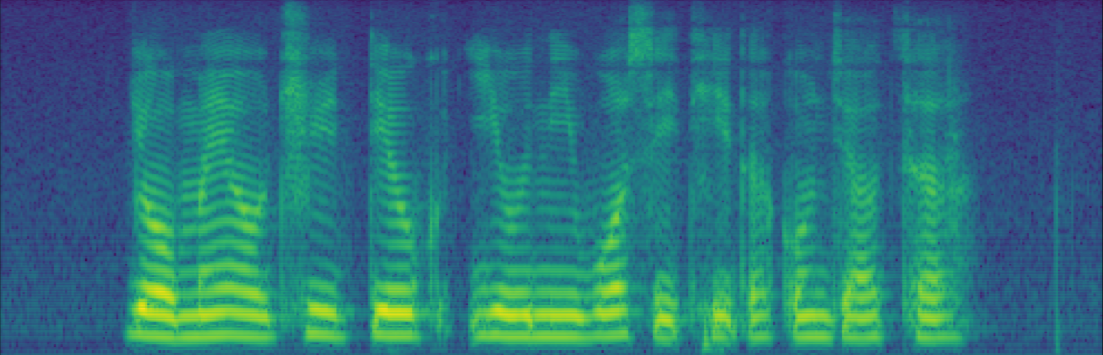}
\label{mel_orig}
}
\subfigure[Audio splicing]{
\includegraphics[width=\linewidth,height=1.5cm]{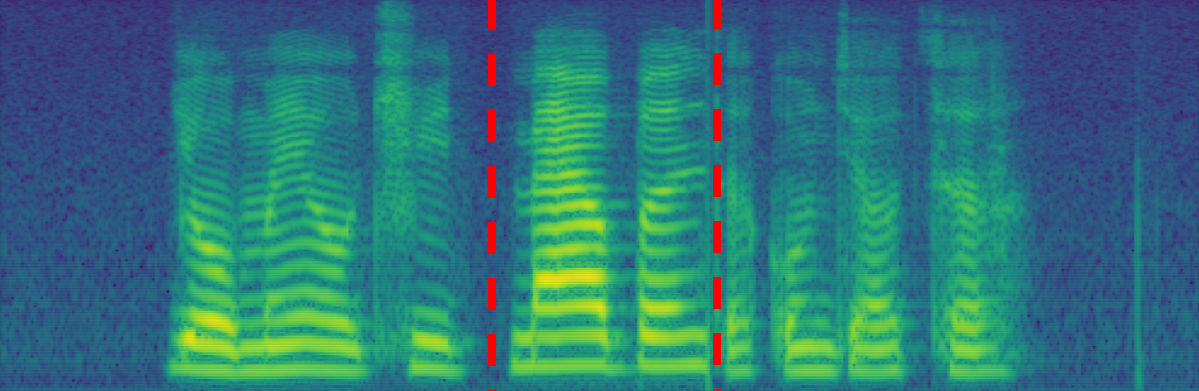}
\label{mel_splic}
}
\subfigure[TTS with word translation]{
\includegraphics[width=\linewidth,height=1.5cm]{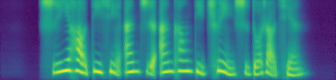}
\label{mel_te}
}
\subfigure[TTS with word insertion]{
\includegraphics[width=\linewidth,height=1.5cm]{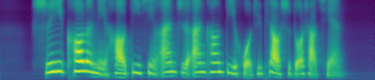}
\label{mel_rie}
}
\caption{Mel spectrograms of the augmented samples}
\label{fig:mels}
\end{figure}

The proposed data augmentation methods for code-switching ASR are evaluated here. The Mel spectrograms of the augmented samples using different methods are shown and compared in Figure \ref{fig:mels}.
For the audio splicing in Figure \ref{mel_splic}, the replaced segment is between the two red lines.

\subsubsection{Audio splicing}
\label{sec: audio_splicing_result}
In order to splice audios, we need alignments between transcripts and speech. We train a GMM-HMM system with a Kaldi \cite{povey2011kaldi} recipe. It starts from a monophone model with MFCC features, finally obtains a tied triphone model. Alignments of ASRU-CS are then calculated with Viterbi beam-search algorithm. According to the alignments, we replace the English segment of each utterance with that of another randomly selected utterance from the same speaker.

\begin{figure}[h]
  \centering
  \includegraphics[width=\linewidth, height=6cm]{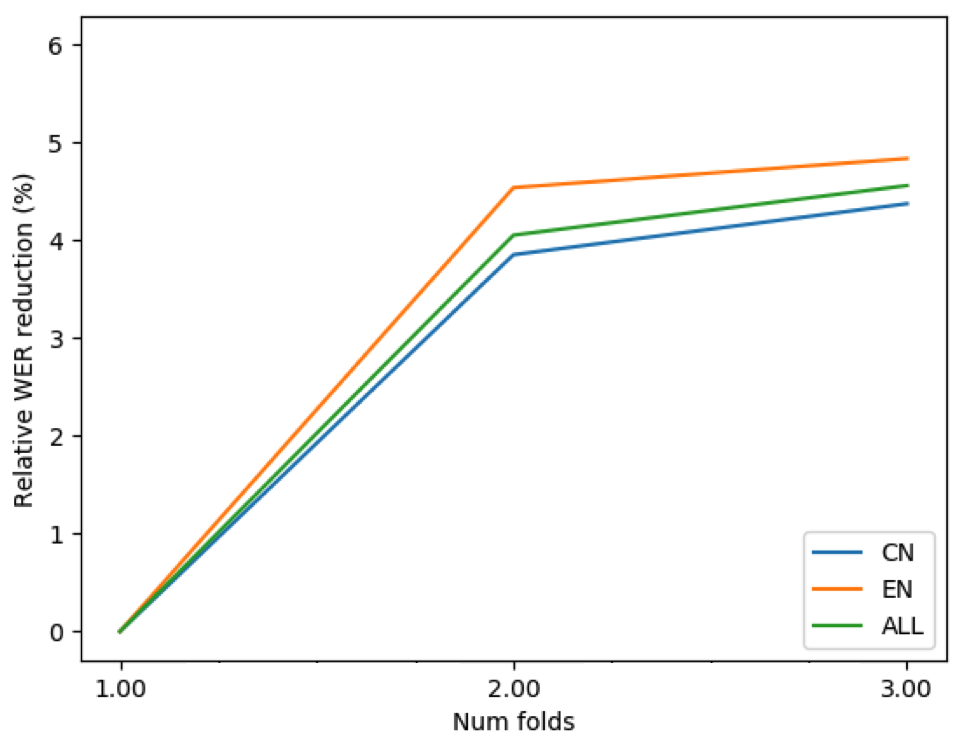}
  \caption{Relation between WER relative reduction and the amount of augmented data using the proposed audio splicing approach.}
  \label{fig:audio splicing}
\end{figure}

We generate a new dataset with the same size as ASRU-CS for data augmentation. The augmented data is used to train a CS E2E ASR system, and the results are shown in Table \ref{tab:cs_aug}. A relative WER reduction of 4.0\% was observed compared to the system without data augmentation, and it is better than the speed perturbation and monolingual TTS methods. We also find that further improvement can be achieved by combining audio splicing with SpecAugment, reducing WER by relative 16.2\% compared to the system without data augmentation, and also relative 4.1\% compared to the system with SpecAugment. 

We also investigate how the amount of additional spliced data affects the performance. Shown in Figure \ref{fig:audio splicing}, 1-fold means the baseline without data augmentation. It shows that we can first get a significant gain with 2-fold data by data augmentation, but only limited improvement when further increasing the augmented data. In all the following experiments, we use 2-fold data for system construction considering both the accuracy and computational cost.

\subsubsection{TTS with word translation}
\label{sec: trans_result}
We apply word translation to 200k ASRU-MANDARIN transcripts for code-switching data augmentation. Specifically, we use Jieba \cite{jieba} toolkit for word segmentation and a bilingual dictionary provided by \cite{choe2020word2word} for Mandarin-to-English word translation. The generated code-switching transcripts is then passed to the TTS system trained on ASRU-CS. The synthetic data and original data are then pooled for end-to-end ASR training.

As shown in Table \ref{tab:cs_aug}, TTS with word translation leads to a relative WER reduction of 5.2\% compared to the baseline result, slightly better than the audio splicing. Compared to the monolingual TTS, synthesizing code-switching data from word translation is more helpful for end-to-end code-switching ASR. Additionally, we find that combining with SpecAugment could obtain a further large improvement.

\begin{table}[t]
\caption{WER (\%) on Mandarin-English test dataset of proposed methods}
\label{tab:cs_aug}
\centering
\begin{tabular}{l|ccc}
\hline
 \multirow{2}{*}{\textbf{Data Augmentation}}    & \multicolumn{3}{c}{\textbf{WER}}  \\
\cline{2-4} 
      & \textbf{CN}            & \textbf{EN}             & \textbf{TOTAL}                                 \\ 
\hline
 None    & 11.15         & 33.31          & 13.56         \\     
\hline
 Audio Splicing        & 10.75   & 31.74  & 13.02 \\ 
 \quad + SpecAug   & 9.23 & 28.81 & 11.36 \\
 Word translation with TTS   & 10.61   & 31.25     & 12.85 \\  
 \quad + SpecAug   & 8.70 & 28.18 & 10.81 \\ 
 
 Word insertion with TTS   & 10.54   & 32.11    & 12.88 \\  
  \quad + SpecAug & 8.51 & 28.17 & 10.65 \\  
\hline
All three proposed + SpecAug & \textbf{8.29} & \textbf{26.74}  & \textbf{10.29} \\ 
\hline
\end{tabular}
\end{table}

\subsubsection{TTS with word insertion}
\label{sec: ins_result}
Finally TTS with word insertion is performed, and we use 200k ASRU-MANDARIN transcripts for this experiment. Specifically, we also apply word segmentation with Jieba \cite{jieba} toolkit and randomly insert an English word from an English lexicon containing words that appear more than 10 times in Librispeech \cite{librispeech} to each Mandarin transcript. Similarly, we pass the generated code-switching transcripts to TTS system and synthesize corresponding speech. Both the synthetic and original dataset are used for end-to-end ASR training. 

Table \ref{tab:cs_aug} shows that the similar observation and the same conclusion as word translation are obtained by TTS with word insertion, and it can also improve the code-switching E2E ASR significantly.

\subsubsection{Combination of all the data-aug methods}

The augmented data from different methods are further combined to increase the diversity of the training data, and the system using all kinds of augmented data is finally constructed. The results are shown as the last line of Table \ref{tab:cs_aug}. It shows that combining different augmented data outperforms all the individual system that uses only a single data augmentation method, indicating that the different data augmentation approaches do not conflict with each other. The final system significantly reduces WER by relative 24.0\% compared to the baseline, and still relative 13.0\% compared to the system with only SpecAugment. 

\section{Conclusions}
\label{sec: conclusion}
In this paper, we propose three new data augmentation approaches for code-switching ASR. We use audio splicing to generate new utterances with only existing utterances, and also attempt to leverage additional monolingual texts by transforming the texts into code-switching ones with word translation or word insertion which is then followed by a TTS system. Experimental results show that all the three approaches yield improvements in code-switching ASR. Additionally, ASR still benefits from our approaches when SpecAugment is further combined. Finally all the different data augmentation approaches are all combined, and the code-switching ASR system can be significantly improved.

\section{Acknowledgements}
This work has been supported by the China NSFC project No. U1736202. Experiments have been carried out on the PI supercomputer at Shanghai Jiao Tong University.

\bibliographystyle{IEEEbib}
\bibliography{main}

\end{document}